\pdfoutput=1

\documentclass[11pt]{article}

\usepackage[final]{acl}

\usepackage{times}
\usepackage{latexsym}
\usepackage[T1]{fontenc}

\usepackage[utf8]{inputenc}
\usepackage{pifont}
\usepackage{microtype}

\usepackage{inconsolata}

\usepackage{graphicx}
\usepackage{natbib}
\usepackage{algorithm, algorithmic}
\usepackage{booktabs}
\usepackage{multirow}
\usepackage{multicol}
\usepackage{enumerate}
\usepackage{amsmath}
\usepackage{xspace} 
\usepackage{xcolor}

\usepackage{times}
\usepackage{latexsym}
\usepackage{natbib}
\usepackage{graphicx}
\usepackage{multirow}
\usepackage{amsmath}
\usepackage{amsfonts}
\usepackage{amssymb,amsfonts}
\usepackage{xcolor}
\usepackage{bbm}
\usepackage{booktabs}
\usepackage{algorithm, algorithmic}
\usepackage{subcaption}
\usepackage{soul}
\usepackage{color, xcolor}
\usepackage{hyperref}
\usepackage{pifont}
\usepackage{lipsum}

\newcommand{\eg}{e.g.\xspace}
\newcommand{\ie}{i.e.\xspace}

\definecolor{mycolor}{RGB}{132,151,176}
\definecolor{ideal}{RGB}{145,172,224}

\title{Pioneering Reliable Assessment in Text-to-Image Knowledge Editing: Leveraging a Fine-Grained Dataset and an Innovative Criterion}

\author{Hengrui Gu\textsuperscript{\ding{71}}, Kaixiong Zhou\textsuperscript{\ding{73}}, Yili Wang\textsuperscript{\ding{71}}, Ruobing Wang\textsuperscript{\ding{71}}, Xin Wang\textsuperscript{\ding{71}\ding{70}} \\
\textsuperscript{\ding{71}}School of Artificial Intelligence, Jilin University \\ 
\textsuperscript{\ding{73}}IMES, Massachusetts Institute of Technology \\
\texttt{ \{guhr22,wangyl21,wangrb22\}@mails.jlu.edu.cn} \\
\texttt{ kz34@mit.edu,xinwang@jlu.edu.cn} \\
}

\begin{document}
\maketitle

\newcommand\blfootnote[1]{%
\begingroup
\renewcommand\thefootnote{}\footnote{#1}%
\addtocounter{footnote}{-1}%
\endgroup
}

\begin{abstract}
During pre-training, the Text-to-Image (T2I) diffusion models encode factual knowledge into their parameters. These parameterized facts enable realistic image generation, but they may become obsolete over time, thereby misrepresenting the current state of the world. Knowledge editing techniques aim to update model knowledge in a targeted way. 
However, facing the dual challenges posed by inadequate editing datasets and unreliable evaluation criterion, the development of T2I knowledge editing encounter difficulties in effectively generalizing injected knowledge. 
In this work, we design a T2I knowledge editing framework by comprehensively spanning on three phases: First, we curate a dataset \textbf{CAKE}, comprising paraphrase and multi-object test, to enable more fine-grained assessment on knowledge generalization. 
Second, we propose a novel criterion, \textbf{adaptive CLIP threshold}, to effectively filter out false successful images under the current criterion and achieve reliable editing evaluation.
Finally, we introduce \textbf{MPE}, a simple but effective approach for T2I knowledge editing. Instead of tuning parameters, MPE precisely recognizes and edits the outdated part of the conditioning text-prompt to accommodate the up-to-date knowledge. A straightforward implementation of MPE (Based on in-context learning) exhibits better overall performance than previous model editors. We hope these efforts can further promote faithful evaluation of T2I knowledge editing methods. Our code is available at 
\href{https://github.com/Hengrui-Gu/T2IKnowledgeEditing}{https://github.com/Hengrui-Gu/T2IKnowledgeEditing}.\blfootnote{\textsuperscript{\ding{70}} Corresponding author}
\end{abstract}


\section{Introduction}
\label{sec:intro}
Text-to-image (T2I) diffusion models have gained significant advancements in encoding real-world concepts via bridging the gap between textual descriptions and visual representations \cite{diff-survey1,diff-survey2,diff-1,diff-2}. By pre-training on a large number of image-caption pairs, these generative models acquire statistical biases on visual concepts such as colors, objects, and personalities. For example, by inputting a text prompt ``the CEO of Tesla", the model can generate a portrait of ``Elon Musk". While some concepts are ageless, other encoded knowledge facts may become invalid over time (\eg, head of a state) or induce harmful social biases (\eg, implicit gender of CEO). To address
this oversight, knowledge editing \cite{gan1,gan2,imageclass,modelediting1,modelediting2,mend,rome,memit,knowledgeediting_add3,knowledgeediting_add1,knowledgeediting_add2} provides an efficient solution by patching undesirable model outputs without significantly altering the model’s general behavior on unrelated input. 



\begin{figure*}[t]
{
\centering
\centerline{\includegraphics[width=\textwidth]{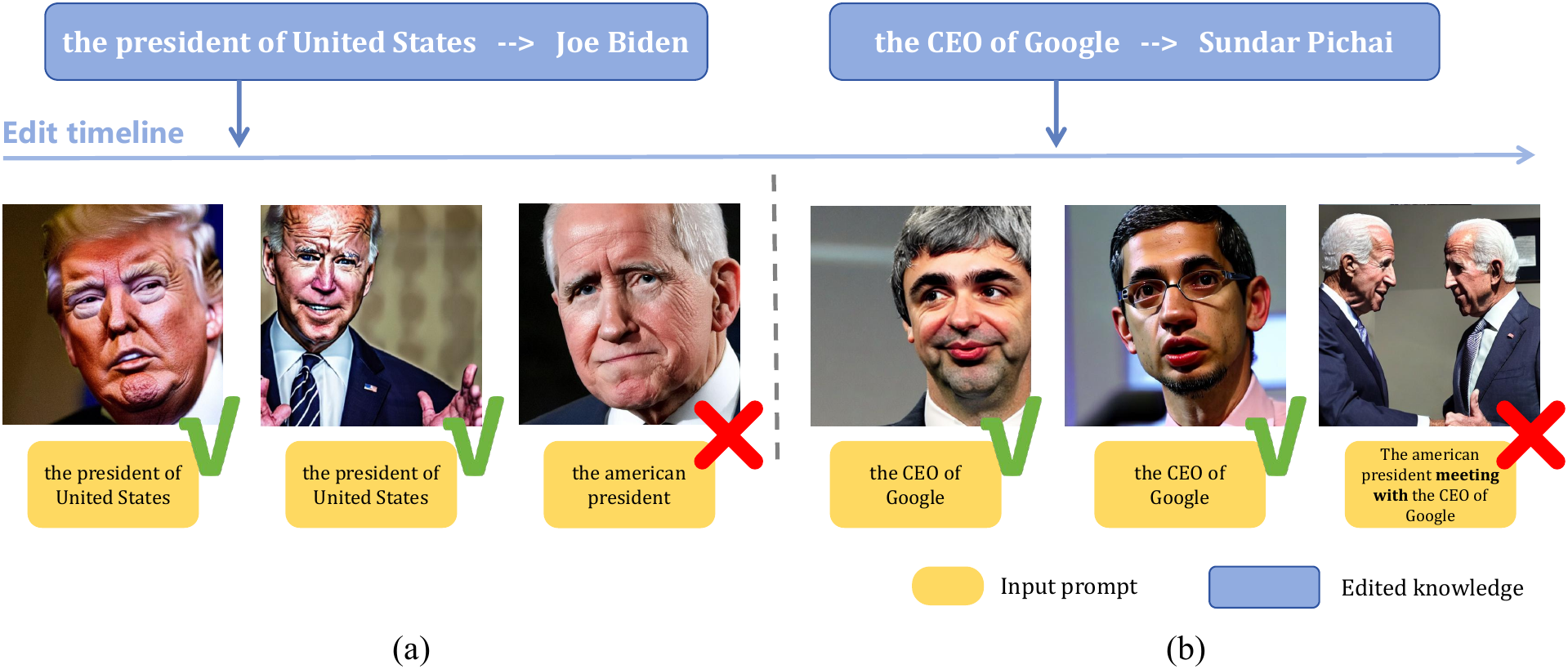}}
\caption{Illustrating the challenges in T2I knowledge editing, the \textbf{timeline} in this figure shows the order in which these images were generated: (a) Existing editing approaches often fail on paraphrases of edit prompt, such as ``the American president''. We term this situation \textbf{Paraphrase Generalization Failure}. (b) The edited model struggles to deal with inputs involved with multiple edited knowledge. We refer to this case as \textbf{Compositionality Generalization Failure}.
}
\label{ill_1}
\vspace{-8pt}
}\end{figure*}

Considering the emerging text-to-image scenario, several pioneering works have been explored for the knowledge editing of generative models~\cite{diff-quickfix,refact,EMCID}. These studies all borrow the idea of localized parameter updating \cite{rome,memit} from language model editing. Specifically, each fact edit is defined as a mapping from edit prompt to target prompt (\eg, "the president of the United States" $\rightarrow$ "Joe Biden") and is represented as a computed key-value vector pair. By locating this vector pair at a specific model component, such as MLP or self-attention block, one is capable of transitioning the generative model's perception on the edit prompt to accord with up-to-date knowledge, thereby achieving knowledge editing.




However, the existing works still focus on exterior model editing, i.e., text mapping, instead of knowledge mapping and generalization reasoning. Based on an edited Stable Diffusion \cite{stablediffusion}, we generate images by creating the input prompts that are synonymous with the fact edit and consist of multiple objects. As illustrated in Fig.~\ref{ill_1}, we observe \ding{192}\textbf{Paraphrase Generalization Failure}: Via replacing the input prompt of fact edit with its paraphrase (\eg, changing "United States" to "American"), the synthetic portrait looks significantly distorted from the ground truth and distinct from the one generated by the original prompt. \ding{193}\textbf{Compositionality Generalization Failure}: When incorporating multiple edited objects within a single input prompt, the model's generation behavior is only partially updated on a subset of fact edits. We attribute these generalization failures to superficial text mapping, where the knowledge editing lacks the reasoning flexibility to adequately comprehend various language concepts.

To effectively address how to implement knowledge mapping in generative models, which requires the edited knowledge to generalize to free and varied language inputs, we must tackle two main challenges. 
\ding{182}Most of the T2I benchmark datasets \cite{time,refact,diff-quickfix} used for knowledge editing do not include complex evaluation prompts comprising paraphrases and multiple edited objects. Such simple datasets hinder the development of sophisticated editing methods associated with the desired generalization capability. 
\ding{183}The evaluation criterion for T2I knowledge editing are underexplored. Namely, given a synthesized image from an edited model, how can we determine whether the synthesis behavior is in line with the desired update?
Previous research \cite{time,refact} formulates the decision of editing success as a binary classification task, comparing the closeness of synthesized images to outdated and target facts. However, as shown in Fig.~\ref{ill_1}, this approach often results in false successful images that appear closer to the target facts but fail to meet the intended editing goals. Thus, a more reliable evaluation strategy is needed to advance knowledge editing efforts.

In response to these challenges, we design a comprehensive text-to-image knowledge editing framework that spans three phases: dataset construction, evaluation strategy, and editing method.
\underline{First}, we curate a dataset named as \underline{C}ounterfactual \underline{A}ssessment of Text-to-image \underline{K}nowledge \underline{E}diting (CAKE) to quantitatively assess the edited model's capabilities in addressing the above-mentioned complex cases. In particular, CAKE introduces two new types of evaluation prompts, built from the paraphrases of edit prompt and multiple edited objects, respectively. In addition to verifying superficial text-mapping, the use of these additional evaluation prompts allows CAKE to offer a more fine-grained assessment of editing performance and insights into how well an editing method generalizes text-mapping to knowledge-mapping.

\underline{Second}, to establish a reliable evaluation strategy for editing, we propose a novel criterion termed adaptive CLIP threshold. Unlike the previous criterion based on classification, this innovative criterion instead focuses on whether the synthesized image is "sufficiently" similar to the target fact. Specifically, this criterion analyzes the CLIP score distribution of ideal synthesized images and utilizes its parameter estimations to calculate a score threshold that quantifies the degree of "sufficiency". Utilizing this score threshold in decision-making can effectively filter out false successful images in editing evaluation scenarios. Our validation experiments supported by \textit{Qwen-vl-max}, the state-of-the-art open-source vision-language model \cite{mmbench,qwen-vl-max} on the celebrity recognition task, demonstrate the superiority of the novel criterion, significantly outperforming the current criterion.





\underline{Third}, rather than tuning parameters, we explore a distinctive approach to T2I knowledge editing termed \textbf{M}emory-based \textbf{P}rompt \textbf{E}diting (MPE). MPE stores all fact edits in an external memory and functions as a pre-processing module for the conditioning text prompt. Before image synthesis, MPE identifies and edits outdated parts of the input prompt to align with current knowledge. Our experiments include a simple, in-context learning-based \citep{icl} implementation of MPE. Extensive results suggest that current editing methods struggle to generalize text-mapping to desired knowledge-mapping, whereas MPE outperforms previous competitors in overall performance and applicability, demonstrating significant potential in addressing T2I knowledge editing.

\section{Related Work}
\textbf{Text-to-image model editing.} Model editing techniques focus on providing stable, targeted updates to model behavior without costly re-training. Related researches have been carried out on a variety of model architectures, such as generative adversarial networks \cite{gan1,gan2}, image classifiers \cite{imageclass} and LLMs \cite{rome,memit,mend,serac}. \citep{time} formally describes T2I model editing as modifying model's generative preference for visual concepts (\eg, editing the default color of \textbf{Roses} from Red to Blue).
 Subsequent studies start to focus on editing factual knowledge in T2I model: Inspiring from language model editing \cite{rome,memit}, ReFACT and Diff-quickfix \citep{refact,diff-quickfix} both encode the to-be-edited knowledge into a key-value vector pair, but place it into different model components (MLP or self-attention block). The concurrent work EMCID \cite{EMCID} sequentially distributes key-value vector pairs across multiple model layers to enable massive concept editing while preserving generation quality. Unlike above methods, our proposed MPE interprets knowledge editing as prompt editing, where the model remains intact, thereby avoiding catastrophic forgetting. This
 
\noindent \textbf{Text-to-image model personalization.}
The goal of T2I model personalization~\citep{personalization1,personalization2,personaliztaion4,personaliztaion5} is to adapt pre-trained T2I models to user-specific image generation needs, enabling high-quality and diverse synthesis of previously unseen visual concepts. This task is fundamentally different from knowledge editing: It focuses on generating images of novel subjects while preserving its class-specific prior~\citep{personalization3}. In contrast, knowledge editing aims to completely rewrite the outdated factual associations within the model, without retaining the originally associated outputs. Since the emergence of visual concepts and factual knowledge updates often occur asynchronously, combining the two techniques together, as outlined in~\citep{refact}, is the most efficient approach to developing practical and flexible T2I models.

\section{Text-to-image Knowledge Editing}

\subsection{Preliminaries}

\label{definition}
\noindent \textbf{Text-to-Image Diffusion Model.} For our analysis, we focus specifically on T2I diffusion models. We consider a T2I diffusion model with deterministic generative processes, as described in \cite{ddim}. This model can be expressed as $f(\textbf{x}_{T}, p)$, where $p$ represents the conditioning text prompt and $\textbf{x}_{T}$ is the initial latent variable sampled from a Gaussian distribution. The function $f$ denotes a deterministic, iterative denoising process, which outputs a real image $\textbf{x}$.


\noindent \textbf{Text-to-Image Knowledge Editing.} Unlike language model editing \cite{rome,mend,mquake,pokemqa}, we define a fact edit $e$ as a text mapping $(p_\mathrm{edit} \rightarrow p_\mathrm{tar})$, for example, (the U.S. president $\rightarrow$ Joe Biden). For practical applicability, we argue that the edited model should generalize the injected edits from external text mappings to internal knowledge mappings. Given an edit $e=(p_\mathrm{edit} \rightarrow p_\mathrm{tar})$, we formally describe the goal of T2I knowledge editing as producing an edited model $f_{\mathrm{edit}}$ based on $f$ and $e$. The edited model $f_{\mathrm{edit}}$ should satisfy the following conditions:

\begin{small}
\begin{equation}\label{goal of KE}
\begin{aligned}
    & \forall p \in \mathrm{Para}(p_{\mathrm{edit}}), & f_{\mathrm{edit}} (\textbf{x}_{T}, p) & = f(\textbf{x}_{T}, p_{\mathrm{tar}}), \\
    & \forall p \notin \mathrm{Para}(p_{\mathrm{edit}}), & f_{\mathrm{edit}}(\textbf{x}_{T}, p) & = f(\textbf{x}_{T}, p),
\end{aligned}
\end{equation}
\end{small}
where $\mathrm{Para}(\cdot)$ represents the set containing all paraphrases of $p_{\mathrm{edit}}$. The objective of this task requires the edited model to recognize $p_\mathrm{edit}$ in any form and map it to $p_\mathrm{tar}$ through the encoding process, which we refer to as knowledge mapping.





\begin{table}[t]
\renewcommand\arraystretch{1.3}
\centering

\resizebox{\linewidth}{!}{
    \footnotesize
\begin{tabular}{l|l}
\hline
\multicolumn{1}{c|}{\textbf{Single}} &
  \textbf{Edit I: the president of the United States -\textgreater Tim Cook} \\ \hline
Efficacy    & \{The president of the United States / Tim Cook\}                     \\ \hline
Generality  & \{The president of the United States / Tim Cook\} in a meeting        \\
            & \{The president of the United States / Tim Cook\} eating an apple  \\ \hline
KgeMap      & \{The leader of the United States / Tim Cook\} runing in the streets  \\
            & \{The U.S. president / Tim Cook\} eating strawberries              \\ \hline
Specificity & \{ flag of the United States / flag of the United States \}                 \\
            & \{ currency of the United States / currency of the United States \}     \\ \hline
\multicolumn{1}{c|}{\textbf{Composite}} &
  \textbf{Edit II: the Titanic male lead -\textgreater Jeff Bezos} \\ \hline
Compo &
  \begin{tabular}[c]{@{}l@{}}\{The president of United States and the Titanic male lead / Tim \\ Cook and Jeff Bezos\} hiking in the mountains\end{tabular} \\
            & \{...\} having a causal conversation at a coffee shop              \\ \hline
\end{tabular}
}
\caption{Part of the first entry in the CAKE dataset. All prompts are represented in $\{p_{\mathrm{edit}}/p_{\mathrm{tar}} \}$. During experiments, each entry undergoes top-down \textbf{alternating} editing for fair comparisons (See Appendix~\ref{sec:dataset} for details), \ie Edit I $\rightarrow$ evaluate \{Efficacy, Generality, KgeMap, Specificity\} $\rightarrow$ Edit II $\rightarrow$ evaluate \{Compo\}.} 
\vspace{-5pt}
\label{CAKE}
\end{table}

\subsection{Counterfactual Assessment of Text-to-image Knowledge Editing}
\label{CAKE_metrics}
In order to faithfully assess how well the editing methods achieve knowledge mapping, we build CAKE (\underline{C}ounterfactual \underline{A}ssessment of Text-to-image \underline{K}nowledge \underline{E}diting) for practical and fine-grained editing evaluation. See Appendix \ref{sec:dataset} for dataset construction process and statistics.


Following previous work (The RoAD dataset, \citealp{refact}), CAKE focus on counterfactual edits about figures associated with specific roles (\eg, editing \textbf{The U.S. president} $\rightarrow$ \textbf{Tim Cook}). This includes a diverse range of roles, such as entrepreneurs, politicians and so on. CAKE totally contains 100 entries and each entry consists of two counterfactual edit prompts and 15 evaluation prompts, which are all represented in the form: $\{p_{\mathrm{edit}}/p_{\mathrm{tar}} \}$, as shown in Table \ref{CAKE}.

\begin{figure*}[t]
{
\centering
\centerline{\includegraphics[width=\textwidth]{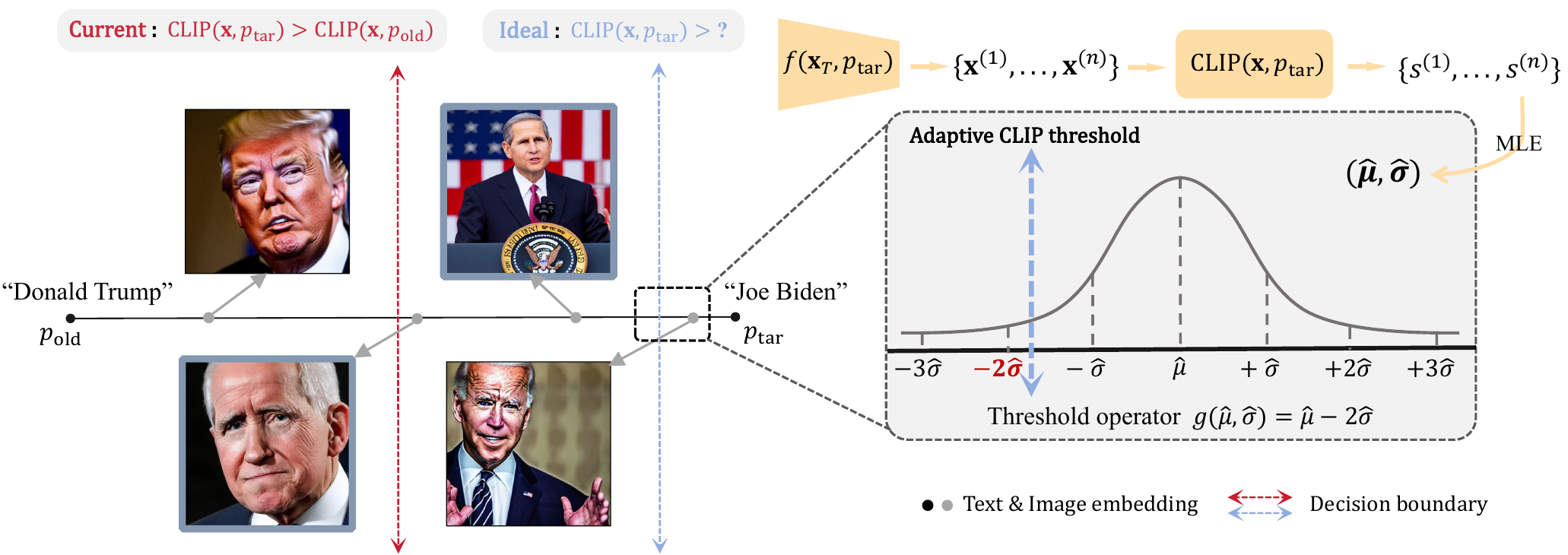}}
\caption{ An editing evaluation example ($p_{\mathrm{edit}}=$"the U.S. president", $p_{\mathrm{tar}}=$"Joe Biden"). A closer distance between two embedding points implies higher similarity, \ie CLIP-Score. The images with \textbf{\textcolor{mycolor}{borders}} are false successful images under the current criterion. For each evaluation prompt, the adaptive CLIP threshold precisely approximates the \textbf{\textcolor{ideal}{ideal decision boundary}} and effectively filters out the false successful images.
}
\label{Ada_metric}
\vspace{-5pt}
}\end{figure*}

After updating the knowledge expressed by the given edit prompts in a T2I model, we use different types of evaluation prompts to compute the editing performance in various dimensions:






\noindent \textbf{Efficacy}: Determine whether the edited model comprehends the updated text mappings.

\noindent \textbf{Generality}: Assess whether the edited model can flexibly utilize the updated text mappings.

\noindent \textbf{Specificity}: Measure how well the edited model preserves other close but unrelated concepts.

\noindent \textbf{KgeMap (New)}: Use paraphrases to verify whether the edited model generalizes updated text mappings to knowledge mappings.

\noindent \textbf{Compo (New)}: Evaluate the edited model's capability to apply multiple updated knowledge elements in its generative behavior simultaneously.

Evaluating in terms of the above fine-grained metrics allows CAKE to serve as a robust starting point for developing more effective and practical editing methods.

\subsection{Adaptive CLIP Threshold Criterion}
\label{ada_clip_threshold}
After updating a fact edit to a T2I model and synthesizing an image conditioned on an evaluation prompt, the critical question becomes: \textbf{How can we determine whether the synthesis aligns with the desired update?}


Previous researches~\cite{refact,time} formulate the question as a binary classification task and use the CLIP-Score $\mathrm{CLIP}(\cdot,\cdot)$~\cite{clip,clipscore} to measure text-image similarity, setting the \textbf{current decision boundary} for determining editing success. However, this approach overlooks whether the synthesized image is "sufficiently" close to the target fact, leading to false positives where ineligible images are mistakenly labeled as successful (see Fig~\ref{Ada_metric}).

To address this, we propose an \textbf{adaptive CLIP threshold} that better aligns with the \textbf{ideal decision boundary}. By analyzing the CLIP-Score distribution of ideal images, we establish a prompt-specific threshold that quantifies "sufficiency", providing a more precise and reliable measure for evaluating edits.




To obtain the threshold, an extra warm-up stage is required before editing, as illustrated in Fig.~\ref{Ada_metric}. For each evaluation prompt $\{p_{\mathrm{edit}}/p_{\mathrm{tar}} \}$, we use the clean T2I model $f$ conditioned on $p_{\mathrm{tar}}$ to generate a set of real images $\left \{ \textbf{x}^{(1)},\dots ,\textbf{x}^{(n)}\right \}$, where $\textbf{x}^{(i)} = f(\textbf{x}_{T}^{(i)},p_{\mathrm{tar}})$ and $\textbf{x}_{T}^{(i)}$ is the randomly sampled initial variable. These real images inherently bear sufficient similarity to the target fact $p_{\mathrm{tar}}$ and are thus considered ideal for post-editing generation, i.e., $f_{\mathrm{edit}}(\textbf{x}_{T},p_{\mathrm{edit}})$.

Next, we calculate the CLIP-Score between these ideal images and $p_{\mathrm{tar}}$ to form an ideal score set $S=\left \{s^{(1)},\dots ,s^{(n)}\right\}$, where $s^{(i)} = \mathrm{CLIP}(\textbf{x}^{(i)},p_{\mathrm{tar}})$. We assume the ideal score $s$ follows a normal distribution $\textit{N}(\mu,\sigma)$ and estimate its parameters $\hat{\mu}$ and $\hat{\sigma}$ using Maximum Likelihood Estimation~\cite{MLE}:

\begin{small}
\begin{equation}\label{parameter estimates}
\hat{\mu} = \frac{1}{n} \sum_{i=1}^{n} s^{(i)}, \quad \hat{\sigma} = \sqrt{\frac{1}{n-1} \sum_{i=1}^{n} (s^{(i)} - \hat{\mu})^2},
\end{equation}
\end{small}
where $\hat{\mu}$ and $\hat{\sigma}$ are the unbiased parameter estimates for $\textit{N}(\mu,\sigma)$. We define an operator $g(\hat{\mu},\hat{\sigma})$ that calculates the minimum successful similarity as the decision-making threshold, to preserve most ideal images while filtering out most unsuccessful images, as follows:
\begin{equation}\label{criterion1}
    \mathrm{CLIP}(f_{\mathrm{edit}}(\textbf{x}_{T},p_{\mathrm{edit}}),p_{\mathrm{tar}}) \geq g(\hat{\mu},\hat{\sigma}).
\end{equation}

\begin{figure}[t]
{
\centering
\centerline{\includegraphics[width=\linewidth]{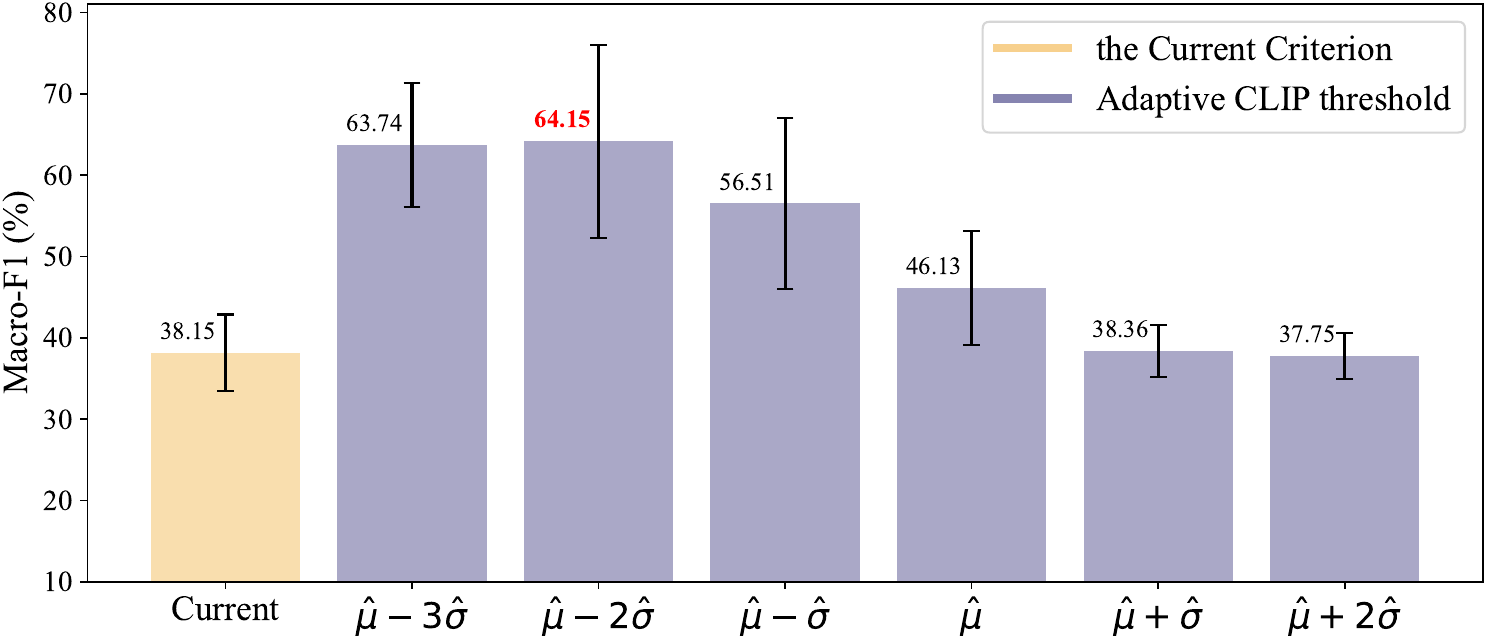}}
\caption{Using \textit{Qwen-vl-max} as the pseudo-label generator, the Macro-F1 performance across different criterion / threshold operators. \textbf{Current} refers to the current, classification-based criterion. }
\label{criterion_validation}
\vspace{-12pt}
}\end{figure}

Eq.~\eqref{criterion1} formulates the new criterion for editing evaluation. To determine the optimal operator $g(\hat{\mu},\hat{\sigma})$ for the knowledge editing task, we conducted a criterion validation experiment. We tested the existing editing method, ReFACT~\cite{refact}, on the role-editing benchmark RoAD~\cite{refact} using several operator choices (e.g., $\hat{\mu}-2\hat{\sigma}$) to make evaluation decisions. Additionally, we selected Qwen-vl-max~\cite{qwen-vl-max}, the best-performing open-source vision-language model for the \textbf{Celebrity Recognition} task \cite{mmbench}, as the pseudo-label generator (see Appendix~\ref{sec:criterion} for the pseudo-label generation process)\footnote{The ability of GPT-4v to perform person identification has been officially prohibited. Thus, Qwen-vl-max was chosen.}. Fig.~\ref{criterion_validation} presents the Macro-F1 performance of various operator choices and the current classification-based criterion. The results demonstrate that $\hat{\mu}-2\hat{\sigma}$ is the most effective choice among the candidate operators\footnote{We also conducted experiments using another VLM , Kosmos-2~\cite{kosmos-2}, for labeling, along with human-annotated labels, which revealed consistent patterns. Detailed results can be found in Appendix~\ref{sec:criterion-validate}}. Furthermore, the adaptive CLIP threshold consistently outperforms the current criterion, indicating its reliability as an evaluation scheme. In later experiments, we set threshold operator $g(\hat{\mu},\hat{\sigma})=\hat{\mu}-2\hat{\sigma}$.


\subsection{MPE: A Proposal for Text-to-Image Knowledge Editing}
\label{MPE}

In this section, we propose a simple and effective scheme for T2I knowledge editing, MPE (\underline{\textbf{M}}emory-based \underline{\textbf{P}}rompt \underline{\textbf{E}}diting). 

\noindent \textbf{Workflow.} Unlike previous parameter-update methods, when receiving a fact edit $(p_\mathrm{edit} \rightarrow p_\mathrm{tar})$, MPE keeps the T2I model frozen and serves as a pre-processing module for the conditioning text prompt $p$, as follows:
\begin{equation}\label{criterion}
    f_{\mathrm{edit}}(\textbf{x}_{T},p)=f(\textbf{x}_{T},\textrm{MPE}(p,p_{\mathrm{edit}},p_{\mathrm{tar}})).
\end{equation}

Towards the task objective defined in Sec \ref{definition}, the expected output of MPE should be either $p_{\mathrm{tar}}$ or $p$, depending on whether $\mathrm{Para}(p_{\mathrm{edit}})$ contains $p$ itself or any sub-sequence of $p$ (\eg, the ideal output of "\underline{The U.S. president} reading a book" should be "\underline{Joe Biden} reading a book").

\begin{figure}[t]
{
\centering
\centerline{\includegraphics[width=\linewidth]{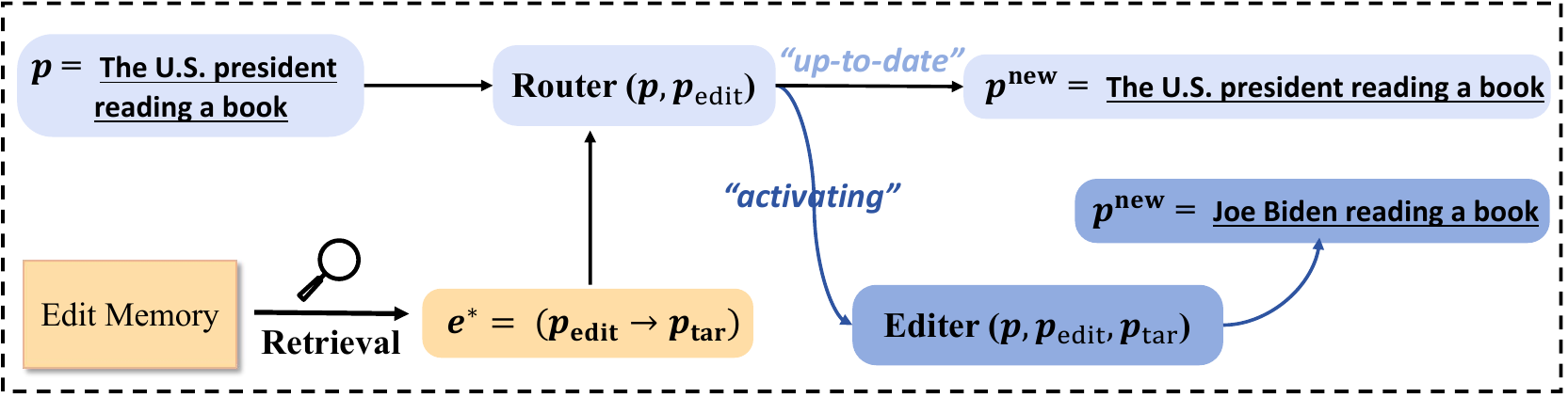}}
\caption{The basic workflow of MPE. }
\label{MPE_workflow}
\vspace{-10pt}
}\end{figure}

In particular, MPE consists of two components: Router and Editer. 1) The Router takes $p$ and $p_{\mathrm{edit}}$ as input and detects whether the $p$ contains any paraphrases from $\mathrm{Para}(p_{\mathrm{edit}})$. If so, it sends an "activating" signal to the Editer, which implies the generating behavior on $p$ of the clean model $f$ has been outdated. 2) If receiving the signal, the Editer would precisely recognize the outdated part (any form of the $p_{\mathrm{edit}}$) of the input prompt $p$ and then replace it with the $p_{\mathrm{tar}}$. Depending on MPE, the text prompt can adaptively fuse with edited knowledge, thereby altering the T2I model's generation behavior in a targeted way, as shown in Fig \ref{MPE_workflow}.

\begin{table*}[tb]
  
  \centering
  \resizebox{\linewidth}{!}{
    \footnotesize
\begin{tabular}{l|cccccc|cc}
\toprule
\multicolumn{1}{c|}{\textbf{Method}} &
  \textbf{Score} &
  \textbf{Efficacy} &
  \textbf{Generality} &
  \textbf{KgeMap} &
  \textbf{Compo} &
  \textbf{Specificity} &
  \textbf{FID $(\downarrow)$} &
  \textbf{CLIP} \\
  \midrule
Base &
  0.00 &
  00.00\%{\scriptsize $\pm$0.00} &
  03.09\%{\scriptsize $\pm$0.93} &
  03.10\%{\scriptsize $\pm$0.67} &
  01.73\%{\scriptsize $\pm$0.66} &
  \textbf{96.90}\%{\scriptsize $\pm$1.53} &
  33.41 &
  0.426 \\
TIME &
  11.4 &
  03.50\%{\scriptsize $\pm$0.92} &
  12.68\%{\scriptsize $\pm$1.73} &
  10.37\%{\scriptsize $\pm$1.62} &
  04.80\%{\scriptsize $\pm$1.17} &
  \underline{85.80}\%{\scriptsize $\pm$3.09} &
  31.94 &
  0.421 \\
{\small ReFACT} &
  35.2 &
  33.70\%{\scriptsize $\pm$6.18} &
  42.46\%{\scriptsize $\pm$5.51} &
  34.10\%{\scriptsize $\pm$4.48} &
  35.73\%{\scriptsize $\pm$4.87} &
  31.19\%{\scriptsize $\pm$2.09} &
  33.38 &
  0.426 \\
{\small EMCID} &
  41.9 &
  82.60\%{\scriptsize $\pm$8.82} &
  48.48\%{\scriptsize $\pm$4.73} &
  39.43\%{\scriptsize $\pm$2.89} &
  40.83\%{\scriptsize $\pm$6.93} &
  19.97\%{\scriptsize $\pm$1.50} &
  32.65 &
  0.426 \\
MPE &
  \textbf{77.2} &
  \textbf{\underline{94.40}}\%{\scriptsize $\pm$2.73} &
  \textbf{\underline{88.84}}\%{\scriptsize $\pm$4.52} &
  \textbf{\underline{63.07}}\%{\scriptsize $\pm$2.52} &
  \textbf{\underline{72.70}}\%{\scriptsize $\pm$3.35} &
  71.20\%{\scriptsize $\pm$1.87} &
  33.41 &
  0.426 \\
  \bottomrule
\end{tabular}

}
\caption{ Quantitative evaluation results on CAKE. Best results are marked with
\textbf{bold}. Best results among editing methods are marked with \underline{underline}. \textbf{FID} refers to FID-5K, \textbf{CLIP} refers to the average CLIP Score.}
\label{main_result}
\vspace{-4pt}
\end{table*}

\noindent \textbf{Multiple editing.} Real-world scenarios generally involve a vast pool of knowledge updates. To operate in practical applications, MPE adopts a "Memory + Retrieval" strategy \citep{serac,pokemqa,black-box} and introduces an additional Retriever component. Specifically, when receiving multiple edits $\left \{e^{(1)},\dots ,e^{(n)}\right \}$, MPE stores all edits in an external memory and embeds their $p_{\mathrm{edit}}^{(i)}$ by the Retriever to construct a retrieval index. Then for each input prompt $p$, the retrieval index returns the key edit $e^{*}$ that is the most relevant (\ie, closest in the embedding space) to $p$, and sends them together to the Router for prompt editing. The complete workflow of MPE is described in Appendix~\ref{sec:mpe_alg}.

\noindent \textbf{Implementation.} The Router and the Editer can be instantiated using various schemes, such as fine-tuning a pre-trained text classification model \citep{distilbert,bert} for the Router and a Seq2Seq model \citep{bart,t5} for the Editer. In this paper, we consider a lightweight, in-context learning-based implementation: We deploy the pre-trained Contriever model \citep{contriever} locally as the Retriever component and teach the GPT-3.5-turbo API \citep{chatgpt} to work as both the Router and the Editer simultaneously, by our manually designed demonstrations (\ie, input-label pairs). The concrete prompts used are detailed in Appendix~\ref{sec:prompts}.

\section{Experiments}
\subsection{Experimental Setup}
In this paper, we investigate both single-editing (updating edits from a single entry at a time) and multiple-editing (updating edits from multiple entries at a time) scenarios for comprehensive assessment. All experiments are conducted using the Stable Diffusion v1-4 model \cite{stablediffusion}.


\noindent \textbf{Dataset.} In addition to the newly constructed CAKE, we include the knowledge editing dataset RoAD \cite{refact} and the preference editing TIME Dataset \citep{time} in our experiments. The TIME Dataset contains 147 variations about visual concepts (\eg, changing the default color of \underline{Roses} from Red to Blue) to assess the performance in editing generative preference.

\noindent \textbf{Baseline.} Except for the unreleased Diff-quickfix \cite{diff-quickfix}, we experiment with all available T2I knowledge editing baselines, including TIME \cite{time}, ReFACT \cite{refact}, and EMCID \cite{EMCID}. TIME targets at modifying generative preferences and cannot be directly applied to RoAD and CAKE due to the incompatible input format. So we implement an adaptation version of TIME that has been empirically demonstrated to be the most effective version in knowledge editing scenarios \citep{refact}. Following prior settings, we include a special case, \underline{Base}, in our single-editing experiments. For each evaluation prompt $\{p_{\mathrm{edit}}/p_{\mathrm{tar}} \}$, \underline{Base} refers to directly inputting $p_{\mathrm{edit}}$ into the unedited model $f$ for generation, serving as a reference baseline.

\begin{table*}[tb]
  \centering
  \resizebox{0.92\linewidth}{!}{
    \footnotesize
\begin{tabular}{c|c|cccc|cc}
\toprule
\textbf{{\scriptsize Dataset}} &
  \textbf{{\scriptsize Method}} &
  \textbf{Score} &
  \textbf{Efficacy} &
  \textbf{Generality} &
  \textbf{Specificity} &
  \textbf{FID$(\downarrow)$} &
  \textbf{CLIP} \\
\midrule
 &
  Base &
  15.8 &
  02.89\%{\scriptsize $\pm$1.66} &
  14.11\%{\scriptsize $\pm$1.10} &
  \textbf{95.98}\%{\scriptsize $\pm$1.26} &
  33.41 &
  0.426 \\
\multirow{2}{*}{{\scriptsize RoAD}} &
  TIME &
  44.6 &
  28.78\%{\scriptsize $\pm$3.12} &
  37.42\%{\scriptsize $\pm$1.59} &
  82.60\%{\scriptsize $\pm$3.39} &
  31.60 &
  0.422 \\
 &
  {\scriptsize ReFACT} &
  57.1 &
  39.11\%{\scriptsize $\pm$4.44} &
  53.53\%{\scriptsize $\pm$2.72} &
  \underline{88.87}\%{\scriptsize $\pm$1.10} &
  33.36 &
  0.426 \\
 &
 {\scriptsize EMCID} &
  78.9 &
  85.00\%{\scriptsize $\pm$4.07} &
  69.18\%{\scriptsize $\pm$3.06} &
  83.51\%{\scriptsize $\pm$1.58} &
  33.09 &
  0.426 \\
 &
  MPE &
  \textbf{\underline{87.6}} &
  \textbf{\underline{90.89}}\%{\scriptsize $\pm$3.58} &
  \textbf{\underline{89.31}}\%{\scriptsize $\pm$2.36} &
  82.69\%{\scriptsize $\pm$1.41} &
  33.41 &
  0.426 \\
\midrule
 &
  Base &
  49.9 &
  25.77\%{\scriptsize $\pm$3.09} &
  50.85\%{\scriptsize $\pm$2.06} &
  \textbf{95.15}\%{\scriptsize $\pm$1.99} &
  33.41 &
  0.426 \\
{\scriptsize TIME} &
  TIME &
  81.8 &
  84.52\%{\scriptsize $\pm$4.46} &
  79.06\%{\scriptsize $\pm$2.43} &
  82.02\%{\scriptsize $\pm$3.34} &
  31.78 &
  0.423 \\
{\scriptsize Dataset} &
  {\scriptsize ReFACT} &
  73.7 &
  65.38\%{\scriptsize $\pm$4.26} &
  70.87\%{\scriptsize $\pm$2.32} &
  \underline{86.31}\%{\scriptsize $\pm$1.36} &
  33.39 &
  0.426 \\
 &
 {\scriptsize EMCID} &
  79.5 &
  88.65\%{\scriptsize $\pm$3.12} &
  80.54\%{\scriptsize $\pm$2.04} &
  70.31\%{\scriptsize $\pm$1.94} &
  33.18 &
  0.426 \\
 &
  MPE &
  \textbf{\underline{86.4}} &
  \textbf{\underline{97.02}}\%{\scriptsize $\pm$1.63} &
  \textbf{\underline{91.58}}\%{\scriptsize $\pm$1.12} &
  72.65\%{\scriptsize $\pm$1.73} &
  33.41 &
  0.426 \\
\bottomrule
\end{tabular}
}
\caption{ Quantitative evaluation results on RoAD and TIME Dataset. Best results are marked with \textbf{bold}. Best results among editing methods are marked with \underline{underline}.}
\label{main_result_timed_road}
\vspace{-5pt}
\end{table*}

\begin{table*}[tb]
  
  \centering
  \resizebox{0.85\linewidth}{!}{
    \footnotesize
\begin{tabular}{c|c|ccccc}
\toprule
\textbf{{\scriptsize Dataset}} &
  \textbf{{\scriptsize Method}} &
  \textbf{\#1} &
  \textbf{\#10} &
  \textbf{\#25} &
  \textbf{\#50} &
  \textbf{\#All} \\
  \midrule
 &
  {\scriptsize TIME} &
  11.36\% &
  00.00\%{\scriptsize (0\%)} &
  00.00\%{\scriptsize (0\%)} &
  00.12\%{\scriptsize (1\%)} &
  00.00\%{\scriptsize (0\%)} \\
\multirow{2}{*}{{\scriptsize CAKE}} &
  {\scriptsize ReFACT} &
  35.24\% &
  27.76\%{\scriptsize (78\%)} &
  23.84\%{\scriptsize (67\%)} &
  21.62\%{\scriptsize (61\%)} &
  20.15\%{\scriptsize (57\%)} \\
 &
  {\scriptsize EMCID} &
  41.87\% &
  33.54\%{\scriptsize (80\%)} &
  30.42\%{\scriptsize (73\%)} &
  29.27\%{\scriptsize (70\%)} &
  25.85\%{\scriptsize (62\%)} \\
 &
  {\scriptsize MPE} &
  \textbf{77.18}\% &
  \textbf{77.17}\%{\scriptsize (99\%)} &
  \textbf{75.54}\%{\scriptsize (97\%)} &
  \textbf{75.93}\%{\scriptsize (98\%)} &
  \textbf{74.83}\%{\scriptsize (96\%)} \\
  \bottomrule
\end{tabular}
}
\caption{ The metric \underline{Score} in multiple editing experiments on CAKE is reported here to characterize the trend in overall editing performance. The \textbf{(\# num)} refers to the size of edit batch. The \textbf{(percent \%)} indicates the percentage to which the editing methods preserve the single-editing performance \textbf{(\# 1)}. Best results are marked with \textbf{bold}. }
\label{multiple_edit_couti_road}
\vspace{-5pt}
\end{table*}

\begin{figure*}[t]
{
\centering
\centerline{\includegraphics[width=\textwidth]{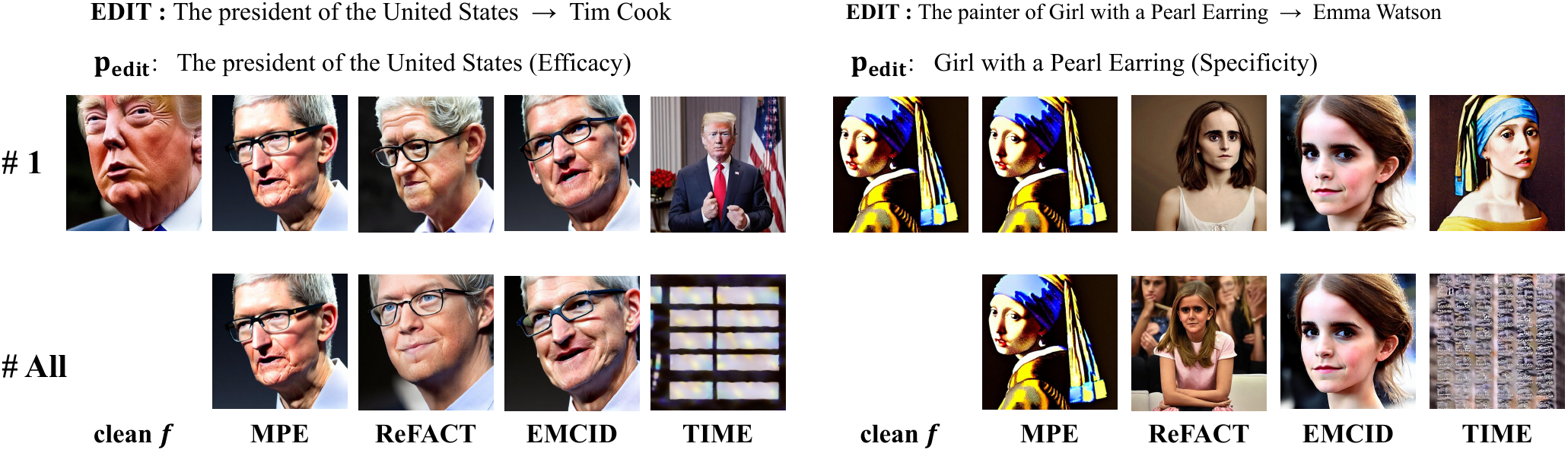}}
\caption{The qualitative examples from the CAKE dataset. The \textbf{(\# num)} refers to the size of edit batch.
}
\label{case_study}
\vspace{-5pt}
}\end{figure*}

\begin{figure*}[t]
    \centering
	\subfloat{
		\includegraphics[width=0.23\linewidth]{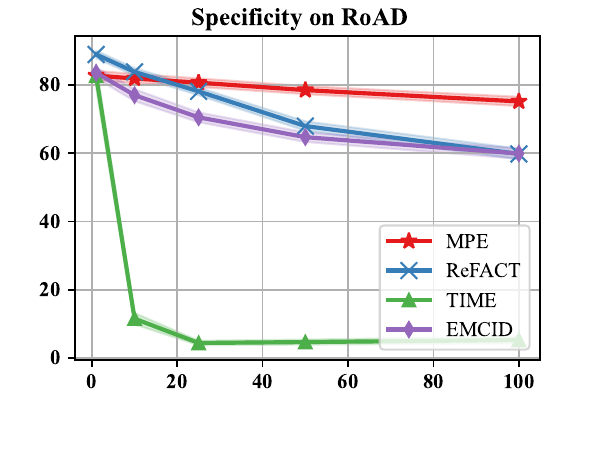}
	}
	\subfloat{
		\includegraphics[width=0.23\linewidth]{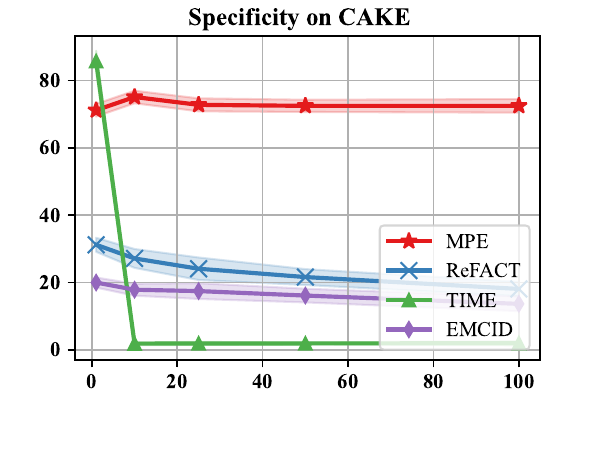}
	}
	\subfloat{
		\includegraphics[width=0.23\linewidth]{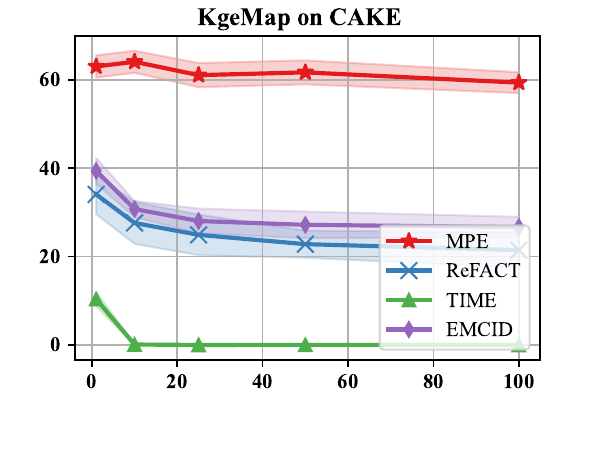}
	}
    \subfloat{
		\includegraphics[width=0.23\linewidth]{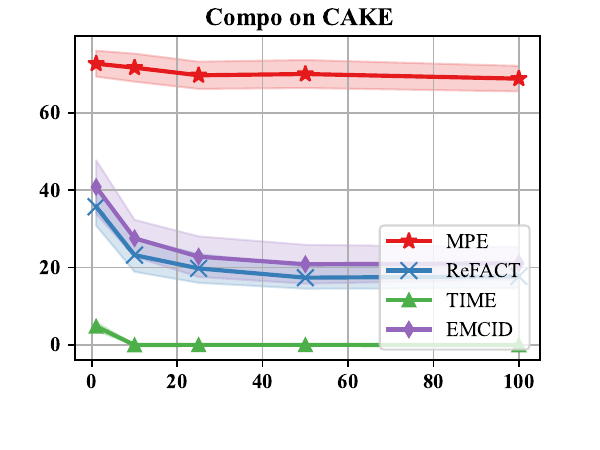}
	}
\caption{The performance curves of various metrics across multiple editing experiments are depicted. The horizontal axis denotes the size of the edit batches, while the shaded areas indicate the standard deviation.}

\label{multiple_figs}
\vspace{-5pt}
\end{figure*}

\noindent \textbf{Metric.} We introduce the metrics we considered in Section \ref{CAKE_metrics}. We evaluate editing performance in terms of Efficacy, Generality, Specificity, KgeMap and Compo. Among them, KgeMap and Compo are only available for the CAKE dataset. We use our proposed adaptive CLIP threshold as the evaluation criterion. After editing, an evaluation prompt $\{p_{\mathrm{edit}}/p_{\mathrm{tar}} \}$ is considered successful if the synthesized image $\textbf{x}$ conditioned on $p_{\mathrm{edit}}$ satisfies $\mathrm{CLIP} (\textbf{x},p_{\mathrm{tar}}) \geq \hat{\mu}-2\hat{\sigma}$. Then each metric is computed as the ratio of successful evaluation prompts to the total number of corresponding evaluation prompts. We also calculate the geometric mean of all the aforementioned metrics as \underline{Score} to characterize the overall performance. To evaluate the general image quality, we report the FID-5K \cite{fid} and the average CLIP score \cite{clip} based on a randomly selected 5,000 image-caption pairs from the MS-COCO validation dataset \cite{coco}. We use Laion’s ViT-G/14 \cite{openclip}, the best open-source CLIP model, to conduct all $\mathrm{CLIP}$ Score calculation.

\noindent \textbf{Setting.} For each evaluation prompt $\{p_{\mathrm{edit}}/p_{\mathrm{tar}} \}$: Before editing, we need an extra warm-up stage to calculate the adaptive CLIP threshold over 50 random seeds; After editing, we generate synthesized images conditioned on $p_{\mathrm{edit}}$ over 10 random seeds to obtain the stable editing performance. Various seeds correspond to different initial variables $\textbf{x}_{T}$. All experiments are conducted on NVIDIA A40s and take about 15 GPU hours to finish one setting.

\subsection{Single Editing Results}

Table \ref{main_result},\ref{main_result_timed_road} presents our single-editing results. We observe that our proposed \textbf{MPE} demonstrates superior overall performance compared to other baselines across all datasets, especially in the knowledge editing task (CAKE, RoAD), underscoring its potential for further development.

The experimental results on CAKE are consistent with our early findings: current editing methods struggle to generalize text-mapping to desired knowledge-mapping, as evidenced by their performance degradation in both the KgeMap and Compo metrics. This poses significant challenges for future research endeavors.



The \textbf{TIME} method, originally designed for editing generative preferences, fails catastrophically on CAKE and thus proves inadequate for updating factual knowledge within the diffusion model. However, its exceptional and well-balanced performance on its initial task (TIME Dataset) remains noteworthy. Considering its low computational cost and rapid editing speed, TIME presents itself as a strong alternative for preference editing.


Quantitatively, the overall performance of \textbf{ReFACT} is relatively low, only surpassing TIME in knowledge editing tasks. Meanwhile, as illustrated by the qualitative examples in Fig.~\ref{case_study}, the synthesis behaviors of the ReFACT-edited model progress in the desired direction but ultimately fail. These "plausible" images can be effectively filtered out using the adaptive CLIP threshold.

\textbf{EMCID} exhibits superior performance among parameter-update editing methods. On RoAD, EMCID distinguishes itself by demonstrating excellent performance across all considered metrics; On CAKE, EMCID is able to generate images that better match the editing goal than ReFACT (See Fig.~\ref{case_study}). However, the weak Specificity in Table~\ref{main_result} indicates that EMCID struggles to limit the editing scope, encountering difficulties in correctly generating close but unrelated concepts after editing.

Interestingly, compared to the superior overall performance, MPE does not excel in Specificity. We attribute this to the drawbacks of prompt editing: once the pre-processing module make a mistake, the revised prompt could be totally unrelated to the original input (\eg, flag of the United States $\rightarrow$ Tim Cook). Fortunately, we later observe that when facing high edit volumes, the Specificity of MPE exhibits excellent robustness, potentially compensating for the identified shortcoming. 





\subsection{Multiple Editing Results}


We conducted multiple editing experiments to simulate real-world scenarios. We group entries into edit batches of size $k$, where $k$ takes values from \{1, 10, 25, 50, all\}. Then for each batch, we injected all fact edits within it into the clean model simultaneously and evaluated the performance on all associated evaluation prompts.

Table~\ref{multiple_edit_couti_road}, Fig.~\ref{multiple_figs} present the related results. We first investigate the changing trend in overall editing
performance: Except MPE, other (parameter-update) editing methods have suffered considerable performance degradation -- TIME completely lost its editing ability; The performance of ReFACT under (\#\textbf{All}) has also declined to nearly half of its single-editing performance; EMCID exhibits better robustness to larger edit volumes, benefited from its distributed editing strategy, but is still significantly inferior to MPE. Utilizing a proficient external retriever, MPE demonstrates outstanding performance retention (96\%) under (\#\textbf{All}). Besides, qualitative examples in Fig.~\ref{case_study} show that 1) TIME frequently generates meaningless pure noise under multiple editing, which reveals the loss in generating ability  caused by parameter updates; 2) ReFACT and EMCID maintain image quality well, suggesting that the MLPs in the text encoder might be a better updating location for knowledge editing.


We then focus on some specific metrics. The curves in Fig.~\ref{multiple_figs} show that MPE owns remarkable robustness to multiple editing, which potentially compensates its weaknesses in Specificity. Conversely, the robustness of ReFACT and EMCID to multiple editing seems less than ideal: They both experience relatively large performance degradation across all metrics. We hope these
results can act as a call to the community to develop more practical and effective editing methods. More quantitative and qualitative results are provided in Appendix~\ref{sec:results}.

\subsection{Performance Analysis on the Retriever Component}
\begin{table}[t]
\renewcommand\arraystretch{1.3}
\centering

\resizebox{0.95\linewidth}{!}{
    \footnotesize
\begin{tabular}{lcccc}
\toprule
Prompt Type & Efficacy & Generality & KgeMap & Compo \\
\midrule
\textbf{\#1}          & 100      & 100        & 100    & 100   \\
\textbf{\#10}         & 100      & 100        & 98.7   & 98.3  \\
\textbf{\#25}         & 100      & 100        & 97.0   & 96.0  \\
\textbf{\#50}         & 100      & 100        & 96.3   & 95.3  \\
\textbf{\#All}        & 100      & 100        & 96.3   & 91.7  \\
\bottomrule
\end{tabular}
}
\caption{\textbf{Retrieval Accuracy} on the CAKE dataset. The \textbf{(\# num)} refers to the size of edit batch. Note that under (\# 1), there is no disturbance edit so the accuracy is always 100. } 
\vspace{-5pt}
\label{Ablation_Retriever}
\end{table}

In this paper, we leverage Contriever~\citep{contriever}, a pre-trained information retrieval model, as the Retriever component of MPE. To validate its reliability and effectiveness for this task, we report its \textbf{Retrieval Accuracy}--the percentage of edit prompts for which the Retriever correctly identifies the relevant edits--in our multiple editing experiments. From the results in Table~\ref{Ablation_Retriever}, we make the following observations:
\begin{itemize}
    \item For the \textit{easy} edit prompt (Efficacy, Generality), Contriever shows no performance degradation under the current editing scale.
    \item For the \textit{hard} edit prompt (KgeMap, Compo), retrieval performance slightly declines as the editing scale increases, indicating that MPE also encounters paraphrase and compositionality challenges. Still, it demonstrates greater robustness compared to parameter-update baselines.
\end{itemize}

\subsection{Time Overhead Analysis of the Adaptive CLIP Threshold}
In Section~\ref{ada_clip_threshold}, we introduce a novel criterion for T2I knowledge editing, referred to as the adaptive CLIP threshold. This approach requires an additional warm-up stage to pre-calculate the decision threshold, which introduces some time overhead.

However, we emphasize that: 1) For each dataset, the warm-up stage is a one-time process, allowing future researchers to bypass this step by using the pre-calculated threshold we provide. 2) Theoretically, this novel criterion reduces evaluation time by half.

\noindent \textbf{Warm-up Stage Time Estimation:} We formally estimate the warm-up time $T_{\mathrm{warm}}$. Below are the notations we define for clarity:
\begin{itemize}
    \item $t_{\mathrm{gen}}$: The time required for a T2I model to generate an image ($t_{\mathrm{gen}} \approx 7.32s$ for Stable-Diffusion v1.4).
    \item $t_{\mathrm{clip}}$: The time required for a CLIP model to compute the CLIP-score once ($t_{\mathrm{clip}} \approx 0.12s$ for Laion's ViT-G/14).
    \item $n_{\mathrm{ideal}}$: The number of ideal images generated to compute the adaptive threshold for each editing entry ($n_{\mathrm{ideal}}=50$ in our experiments).
    \item $n$: The number of editing entries in the dataset ($n=1500$ for the CAKE dataset).
\end{itemize}

The warm-up time can be approximated by the formula $T_{\mathrm{warm}}=(t_{\mathrm{gen}}+t_{\mathrm{clip}})\cdot n_{\mathrm{ideal}} \cdot n$, which gives $T_{\mathrm{warm}}=(7.32s+0.12s)\cdot 50 \cdot 1500 = 155\mathrm{hours}$.

\noindent \textbf{Evaluation Time Estimation:} During evaluation, if we generate $n_{\mathrm{seed}}$ images for each editing entry, the current criterion requires approximately $2*t_{\mathrm{clip}}*n_{\mathrm{seed}}*n$ while the adaptive CLIP threshold requires about $t_{\mathrm{clip}}*n_{\mathrm{seed}}*n$. Hence, the novel criterion reduces the evaluation time by half.


  

\section{Conclusion}

In this work, we aim to establish a reliable evaluation paradigm for T2I knowledge editing. Specifically, we curate a dataset named CAKE, comprising fine-grained metrics to validate knowledge generalization. We then develop an innovative criterion, the adaptive CLIP threshold, to approximate the ideal decision boundary, effectively filtering out false successful images in evaluation scenarios. Additionally, by transferring the editing impact from the parameter space to the input space, we design a distinctive approach, MPE, to achieve T2I knowledge editing. Extensive results have  demonstrated the limitations of current editing methods and the further potential of MPE.

\section*{Limitations}
\label{limitataions}
 The limitations of our work are as follows:

\begin{enumerate}
    \item Similar to previous datasets, our curated CAKE focuses on figure editing pertaining to specific roles. To maintain the quality of evaluation prompts, the scale of CAKE is kept small, comprising only 100 edits and 1,500 evaluation prompts. We suggest that future research should aim to construct a larger and more diverse knowledge editing dataset to achieve more reliable evaluations.
    
    \item Our experiments only involve a straightforward, API-based implementation of our proposed MPE. The further potential of MPE in real applications is under-explored because the call of OpenAI API leads to inevitable financial costs. In future work, we will experiment with more economical schemes of MPE as stated in Sec.~\ref{MPE}.
    
    \item Memory-based editing allows for lossless editing of models and thus distinguishes itself among editing techniques. However, its vulnerability to attacks such as memory injection poses significant risks in production environments. Therefore, this approach requires robust security measures to mitigate these risks effectively in real-world scenarios.
\end{enumerate}


\section*{Ethics Statement}
We curate a counterfactual editing dataset named CAKE, which includes world-renowned roles and identifiable figures. During the dataset construction process, we faithfully adhere to privacy regulations and collect publicly available information from the internet. We randomly assign counterfactual relations between specific roles and figures. On behalf of all authors, we declare that these counterfactual relations are exclusively intended for research purposes and carry no implications for the real world. We have manually ensured that the finished dataset does not contain any potentially offensive content.

\bibliography{custom}

\appendix

\section{Statistics and Construction Details of CAKE}
\label{sec:dataset}

\textbf{Statistics.} CAKE comprises 100 different edits and 1,500 evaluation prompts. Each entry includes two edits (\textbf{Edit I}, \textbf{Edit II}) along with the corresponding evaluation prompts for performance assessment: 1 Efficacy prompt, 5 Generality prompts, 3 Specificity prompts, 3 KgeMap prompts, 3 Compo prompts.

\noindent \textbf{Construction Details.} Given the powerful text generation capabilities of LLMs \cite{llmsurvey1,llmsurvey2,add1,add2,add3}, we utilize ChatGPT to automatically gather candidate edit prompts $p_{\mathrm{edit}} $ and target prompts $p_{\mathrm{tar}}$ to form fact edits. Specifically, we prompt ChatGPT to:
\begin{enumerate}[i)]
    \item list the top-20 influential individuals across various fields of our time (\eg, Jeff Bezos, Tim Cook) to create a candidate target set $\mathcal{O} = \left \{ p_{\mathrm{tar}}^{(1)},\dots ,p_{\mathrm{tar}}^{(20)}\right \}$. We manually verified their correct generation of Stable Diffusion v1-4 \cite{stablediffusion}, the text-to-image diffusion model we study. 
    \item generate 10 roles in different categories (\eg, the CEO of Microsoft).
    \item for each role, leverage in-context learning \cite{icl} to automatically produce 9 additional roles in same category (\eg, the CEO of Tesla, the CEO of IBM) to gather a candidate edit prompt set $\left \{ p_{\mathrm{edit}}^{(1)},\dots ,p_{\mathrm{edit}}^{(100)}\right \}$.
\end{enumerate}

Then for each existing $p_{\mathrm{edit}}$, we randomly assign a target prompt in $\mathcal{O}$ to it and construct a counterfactual text-mapping (edit) set $\mathcal{E}=\left \{e_{1},\dots ,e_{100} \right \}$. We refer to each existing edit as \textbf{Edit I} and build evaluation prompts for them to compose the complete entry. In particular, for all metrics except Specificity, we fill the $p_{\mathrm{edit}}$/$p_{\mathrm{tar}}$ pairs into natural language templates (\eg, \_ eating an apple) to form evaluation prompts. In the case of Specificity, we manually design evaluation prompts (\eg, Tesla logo) inquiring about other knowledge related to the entities (\eg, Tesla) in $p_{\mathrm{edit}}$. 

We then further augment the existing dataset by introducing \textbf{Edit II}: For each entry, we supplement it with a randomly sampled edit $(p_\mathrm{edit}^{\prime} \rightarrow p_\mathrm{tar}^{\prime})$ from the rest of single-edit part that satisfies $p_{\mathrm{tar}} \neq p_{\mathrm{tar}}^{\prime}$. We term the newer edit as \textbf{Edit II}. 

Finally, each candidate entries was independently reviewed by us in terms of grammar and semantic logic. The outcome of this meticulous process was the CAKE dataset comprising 100 entries.

\noindent \textbf{The top-down alternating editing.} The editing and evaluation order of CAKE is slightly different from other editing datasets. After updating the \textbf{Edit I} to the T2I model, we first finish the generations on evaluation prompts of \{ Efficacy, Generality, Specificity, KgeMap\}. Afterwards, we directly insert the \textbf{Edit II} into the current, edited model and finally compute the last metric \{ Compo\}. By following the top-down alternating editing, we test the Compositionality property and can precisely compute the editing performance of T2I model with only one newer edit, aligning with other editing datasets.

\section{Detailed process of the Criterion Validation Experiments}
\label{sec:criterion}
To validate the superiority of our proposed adaptive CLIP threshold and determine the most suitable, human-aligned threshold operator, we leverage two powerful vision-language models(VLMs), \textit{Qwen-vl-max}~\citep{qwen-vl-max} and \textit{Kosmos-2}~\citep{kosmos-2} as the pseudo-label generator, enabling automatic criterion evaluation. Given the excellent performance of current VLMs, it is widely acknowledged that VLM-based automatic validation is fairly reliable and more reproducible than human evaluation~\citep{vlm-eva}.

Specifically, we first apply the T2I knowledge editing method, \textbf{ReFACT}~\citep{refact}, to synthesize edited images for the \textbf{Efficacy} prompt corresponding to each role-editing entry in RoAD, the knowledge editing dataset. Next, we utilize predefined prompts to instruct the VLMs to perform the \textbf{celebrity recognition} task on these edited images in a zero-shot manner: Given an edited image as visual input, the VLMs are prompted to answer the question, `Who is this person?' by choosing from four specific role options and a \textit{None of the above} option. One of the role options corresponds to the target figure after editing, while the others are randomly selected from a pool of candidates of the same gender as the target figure. A synthesized image is labeled as 'successful' only if the VLMs select the correct option or directly output the name of the target figure.


   
    


\section{Additional Results from the Criterion Validation Experiments.}
\label{sec:criterion-validate}

\begin{figure}[t]
{
\centering
\centerline{\includegraphics[width=\linewidth]{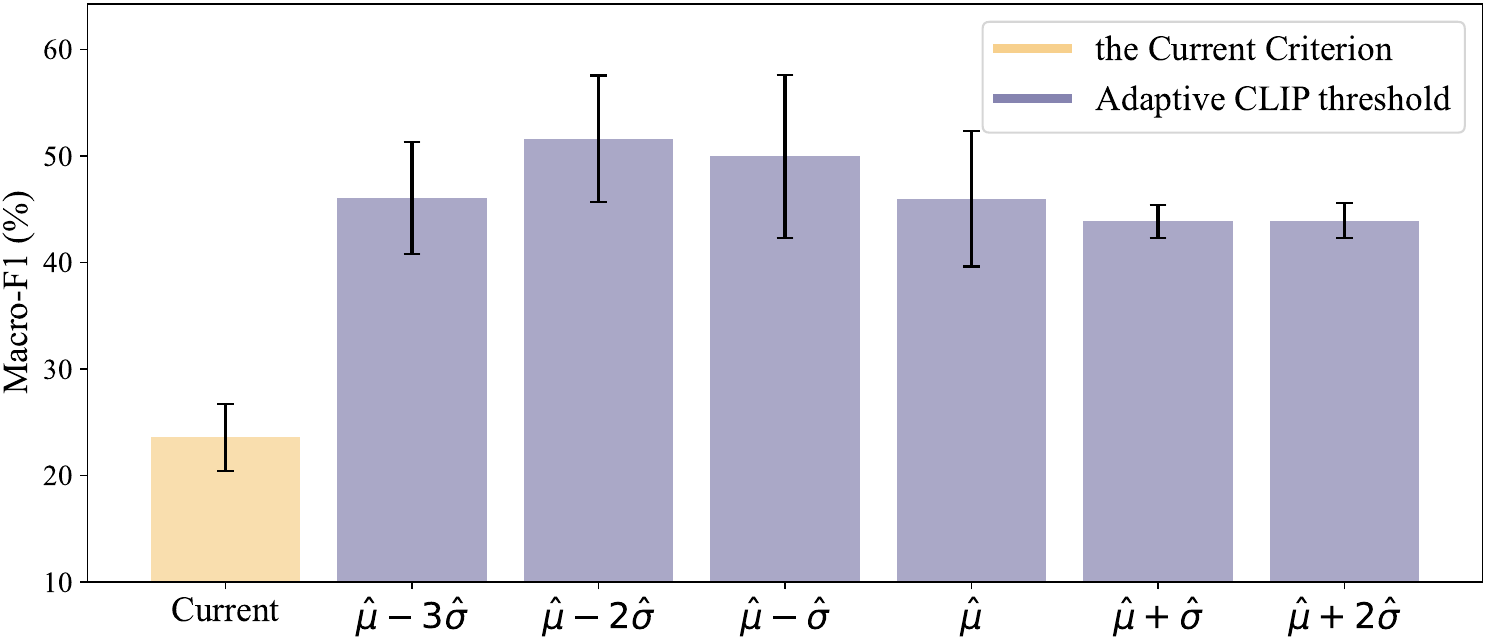}}
\caption{Using \textit{Kosmos-2} as the pseudo-label generator, the Macro-F1 performance across different criterion / threshold operators. \textbf{Current} refers to the current, classification-based criterion. }
\label{criterion_validation_kosmos-2}
\vspace{-12pt}
}\end{figure}

\begin{figure}[t]
{
\centering
\centerline{\includegraphics[width=\linewidth]{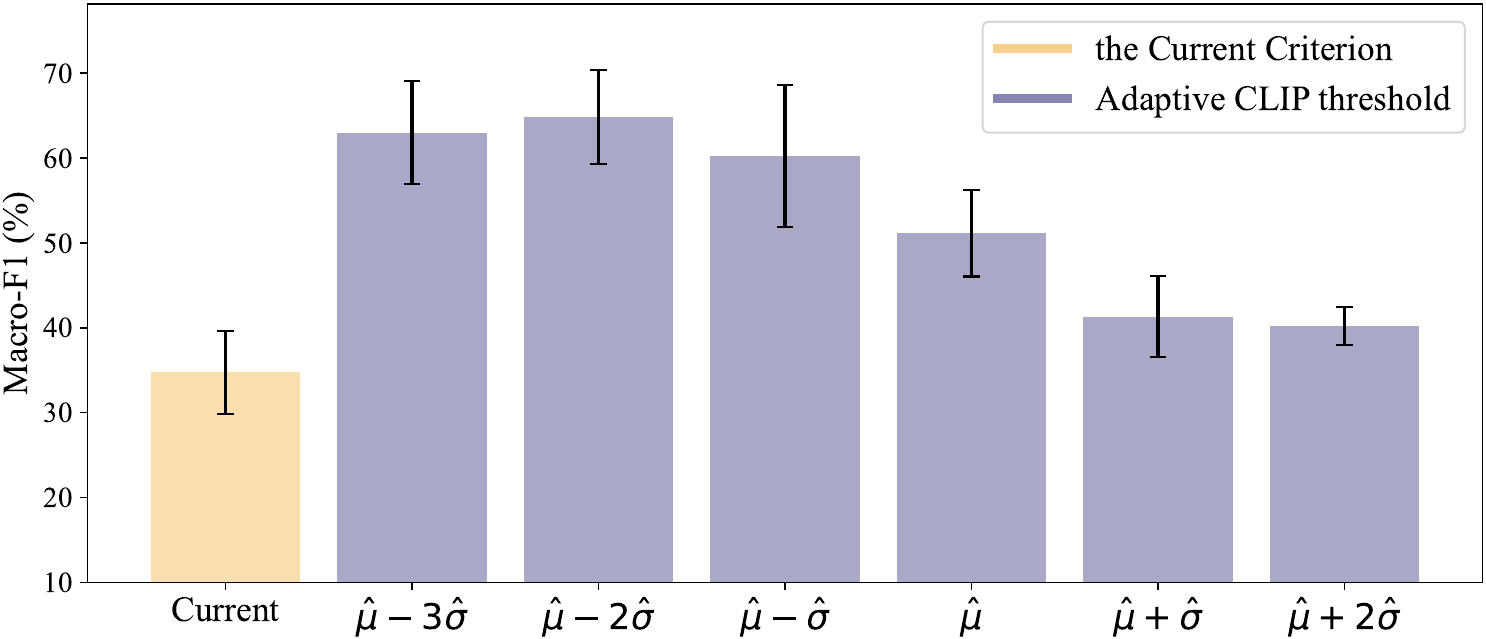}}
\caption{Using \textit{human-annotated} labels, the Macro-F1 performance across different criterion / threshold operators. \textbf{Current} refers to the current, classification-based criterion. }
\label{criterion_validation_humanevaluation}
\vspace{-12pt}
}\end{figure}

Additional results of criterion validation experiments using \textit{Kosmos-2} as label generator and \textit{human-annotated labels} are presented in Fig.~\ref{criterion_validation_kosmos-2} and Fig.~\ref{criterion_validation_humanevaluation}, respectively.

\section{Overall Algorithm of MPE}
\label{sec:mpe_alg}

In Sec~\ref{MPE}, we present the basic workflow of MPE. However, in real applications, when receiving a text prompt $p$, we don't actually know how many fact edits it's associated with. So, to accommodate this problem, we leverage the Router $R$ to determine whether the editing process should be terminated. The specific algorithm is in Alg.~\ref{alg:1}.
\begin{algorithm}[ht]
	\renewcommand{\algorithmicrequire}{\textbf{Input:}}
	\renewcommand{\algorithmicensure}{\textbf{Output:}}
	\caption{Overall Workflow of MPE.}
	\label{alg:1}
	\begin{algorithmic}[1]
		\REQUIRE edit memory $ \mathcal{M}= \{ {e^{(1)} ,\dots , e^{(n)}} \}$, router $R$, editer $E$, retriever $\mathrm{Retrieval}()$, input text prompt $p$
        
        \STATE {\color{blue}/* Editing in the loop */}
        \FOR{$ \mathcal{M} \neq \varnothing $}
        \STATE $e^{*} = \mathrm{Retrieval}(\mathcal{M},p) $
        \STATE $\mathcal{M}=\mathcal{M} \setminus \{e^{*}\}$
        
        \IF{$ R(p,p_{\mathrm{edit}}^{*}) \neq $ "Activating"}
        \STATE \textbf{return} $p$
        \ENDIF
        \STATE $ p = E(p,p_{\mathrm{edit}}^{*},p_{\mathrm{tar}}^{*}) $
		\ENDFOR
	\end{algorithmic} 
\end{algorithm}

\section{Prompts used for In-context Learning}
\label{sec:prompts}
We present several demonstrations from MPE's in-context prompt in Table~\ref{app:prompts} to illustrate the working mechanism of in-context learning-based MPE implementation.

\begin{table}[t]
  
  \centering
  \resizebox{\linewidth}{!}{
    \footnotesize
\begin{tabular}{lll}
\toprule
\multicolumn{3}{c}{\textbf{Several demonstrations in MPE's in-context prompt}}             \\
\midrule
\multicolumn{3}{l}{\textbf{1.}}                                                            \\
\multicolumn{3}{l}{Input: The spokesman of United Nations giving a speech}                 \\
\multicolumn{3}{l}{source concept: The chief trainer of Inter Miami.}                      \\
\multicolumn{3}{l}{target concept: David Beckham.}                                         \\
\multicolumn{3}{l}{Does the entity specified by source concept appeared in the Input: No.} \\
\multicolumn{3}{l}{Output: The spokesman of United Nations giving a speech}                \\
\midrule
\multicolumn{3}{l}{\textbf{2.}}                                                            \\
\multicolumn{3}{l}{Input: The lead singer of Nightwish standing on the stage}              \\
\multicolumn{3}{l}{source concept: The lead singer of Nightwish.}                          \\
\multicolumn{3}{l}{target concept: Elvis Presley.}                                         \\
\multicolumn{3}{l}{Does the entity specified by source concept appeared in the Input: Yes.} \\
\multicolumn{3}{l}{Output: Elvis Presley standing on the stage}                            \\
\midrule
\multicolumn{3}{l}{\textbf{3.}}                                                            \\
\multicolumn{3}{l}{Input: Kylian Mbappe and Kanye West celebrating Christmas together}     \\
\multicolumn{3}{l}{source concept: The chief scientist at NASA.}                           \\
\multicolumn{3}{l}{target concept: Boris Johnson.}                                         \\
\multicolumn{3}{l}{Does the entity specified by source concept appeared in the Input: No.} \\
\multicolumn{3}{l}{Output: Kylian Mbappe and Kanye West celebrating Christmas together}   \\
\bottomrule
\end{tabular}

}
\caption{ Here are several demonstrations from MPE's in-context prompt. When the language model answers the question, 'Does the entity specified by the source concept appear in the input?', it functions as the Router. When the language model generates the final output, it functions as the Editer.}
\label{app:prompts}
\vspace{-5pt}
\end{table}


\section{More Quantitative and Qualitative Results}
\label{sec:results}

\begin{figure*}[t]
    \centering
	\subfloat{
		\includegraphics[width=0.23\linewidth]{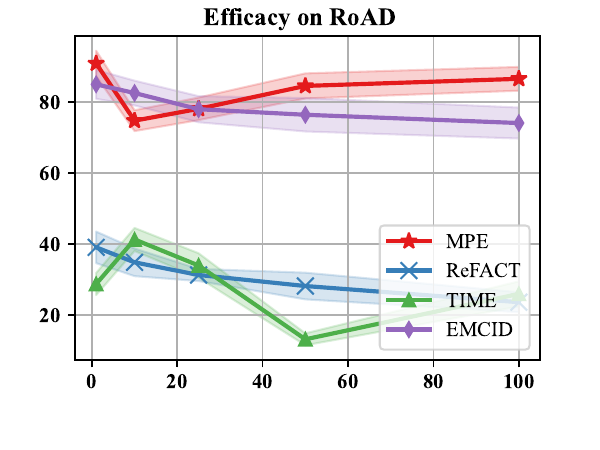}
	}
	\subfloat{
		\includegraphics[width=0.23\linewidth]{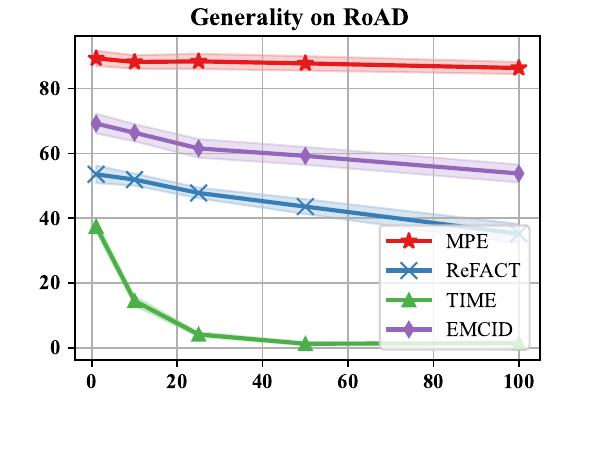}
	}
	\subfloat{
		\includegraphics[width=0.23\linewidth]{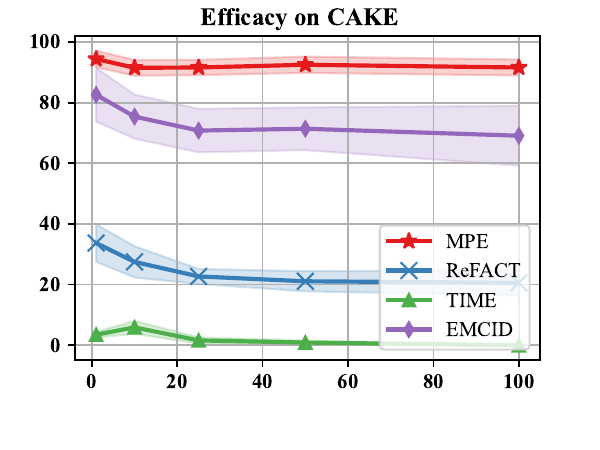}
	}
    \subfloat{
		\includegraphics[width=0.23\linewidth]{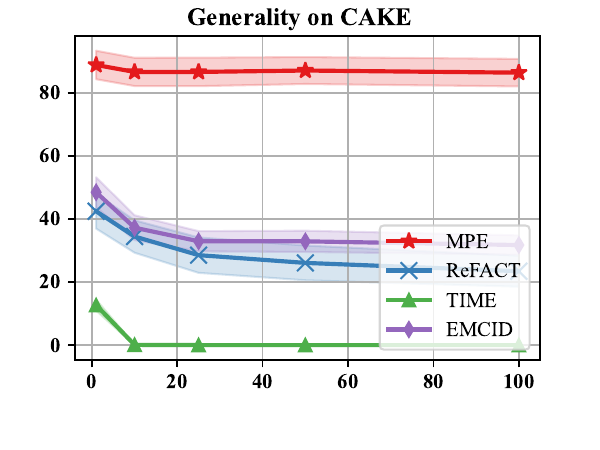}
	}
\caption{The performance curves of various metrics across multiple editing experiments are depicted. The horizontal axis denotes the size of the edit batches, while the shaded areas indicate the standard deviation.}

\label{app:figs}
\vspace{-5pt}
\end{figure*}

The performance curves of editing methods in terms of \{ Efficacy, Generality\} are presented in Fig.~\ref{app:figs}.

The results of the metric \underline{Score} on RoAD in multiple-editing are shown in Table~\ref{app:multiple}.

\begin{figure*}[t]
{
\centering
\centerline{\includegraphics[width=\textwidth]{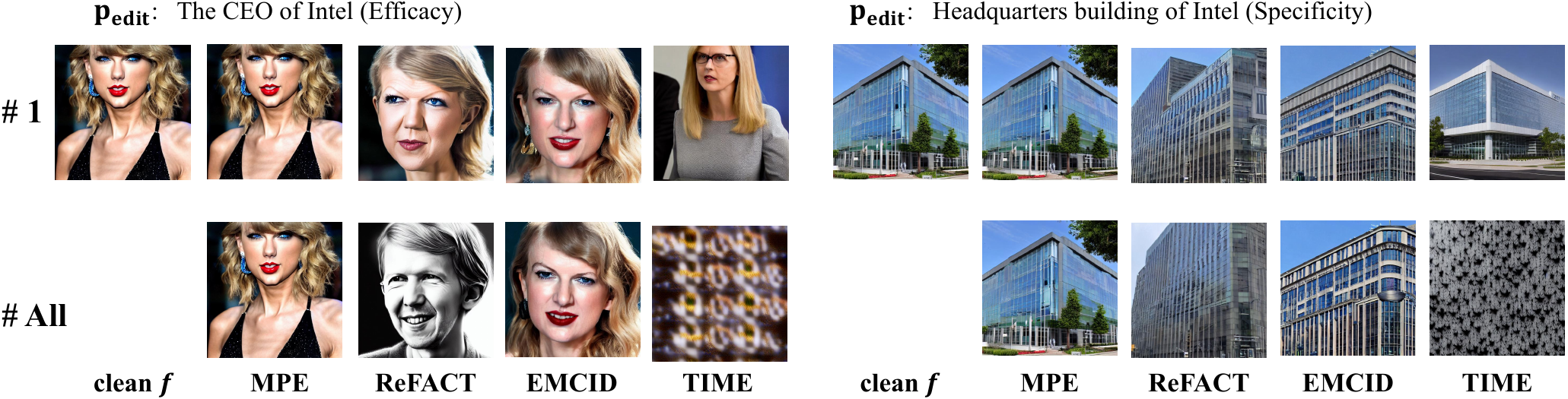}}
\caption{The qualitative examples from the CAKE dataset. The \textbf{(\# num)} refers to the size of edit batch.
}
\label{app:case}
\vspace{-5pt}
}\end{figure*}

Additional qualitative examples in metrics \{ KgeMap, Compo \} are provided in Fig.~\ref{app:case}

\begin{table*}[tb]
  
  \centering
  \resizebox{0.9\linewidth}{!}{
    \footnotesize
\begin{tabular}{c|c|ccccc}
\toprule
\textbf{{\scriptsize Dataset}} &
  \textbf{{\scriptsize Method}} &
  \textbf{\#1} &
  \textbf{\#10} &
  \textbf{\#25} &
  \textbf{\#50} &
  \textbf{\#All} \\
  \midrule
 &
  {\scriptsize TIME} &
  44.64\% &
  19.03\%{\scriptsize (42\%)} &
  8.52\%{\scriptsize (19\%)} &
  04.25\%{\scriptsize (9\%)} &
  05.80\%{\scriptsize (12\%)} \\
\multirow{2}{*}{{\scriptsize RoAD}} &
  {\scriptsize ReFACT} &
  57.09\% &
  53.33\%{\scriptsize (93\%)} &
  48.89\%{\scriptsize (85\%)} &
  43.70\%{\scriptsize (76\%)} &
  36.78\%{\scriptsize (64\%)} \\
 &
  {\scriptsize EMCID} &
  78.89\% &
  74.99\%{\scriptsize (95\%)} &
  69.67\%{\scriptsize (88\%)} &
  66.40\%{\scriptsize (84\%)} &
  62.03\%{\scriptsize (78\%)} \\
 &
  {\scriptsize MPE} &
  \textbf{87.56}\% &
  \textbf{81.42}\%{\scriptsize (92\%)} &
  \textbf{82.26}\%{\scriptsize (93\%)} &
  \textbf{83.50}\%{\scriptsize (95\%)} &
  \textbf{82.49}\%{\scriptsize (94\%)} \\
  \bottomrule
\end{tabular}
}
\caption{ The metric \underline{Score} in multiple editing experiments on RoAD is reported here to characterize the trend in overall editing performance. The \textbf{(\# num)} refers to the size of edit batch. The \textbf{(percent \%)} indicates the percentage to which the editing methods preserve the single-editing performance \textbf{(\# 1)}. Best results are marked with \textbf{bold}.}
\label{app:multiple}
\vspace{-5pt}
\end{table*}

\begin{figure*}[t]
{
\centering
\centerline{\includegraphics[width=\textwidth]{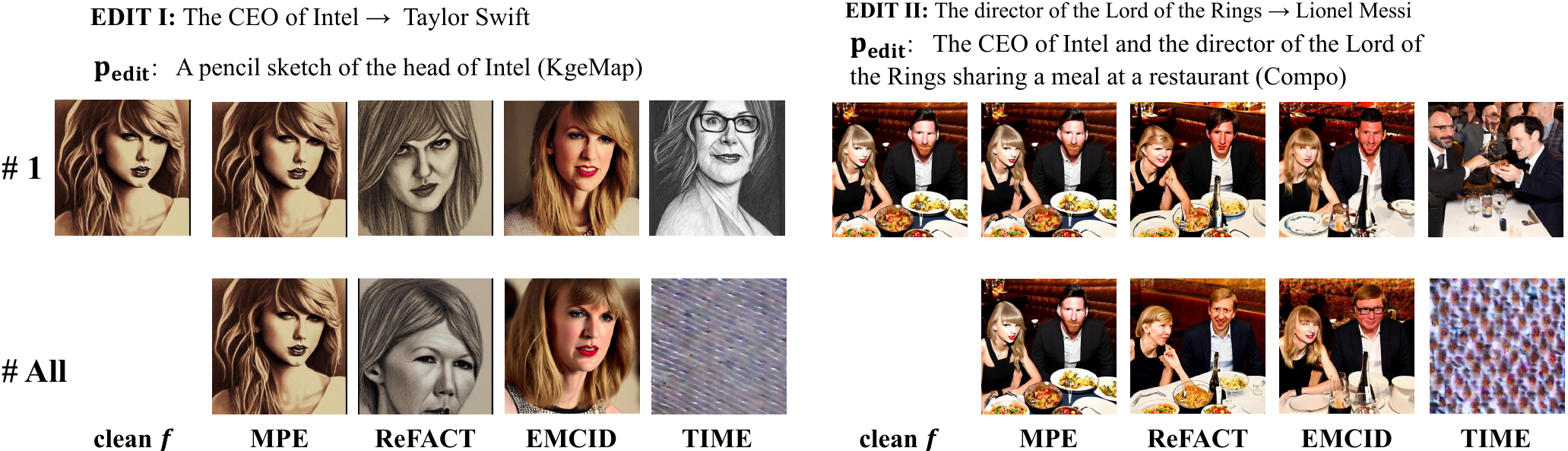}}

\vspace{-5pt}
}\end{figure*}

\end{document}